\pdfoutput=1

\documentclass[11pt]{article}
\usepackage{times}
\usepackage{latexsym}
\usepackage[T1]{fontenc}
\usepackage[utf8]{inputenc}
\usepackage{microtype}
\usepackage{inconsolata}
\usepackage{graphicx}
\usepackage{CJKutf8}
\usepackage{multirow}
\usepackage{booktabs}
\usepackage{amssymb}
\usepackage{amsmath}
\usepackage{algorithm}
\usepackage{algorithmic}
\usepackage{appendix}
\usepackage{stfloats}
\usepackage[final]{coling}

\usepackage{times}
\usepackage{latexsym}

\usepackage[T1]{fontenc}

\usepackage[utf8]{inputenc}

\usepackage{microtype}

\usepackage{inconsolata}

\usepackage{graphicx}

\title{DragFT: Adapting Large Language Models with Dictionary and Retrieval Augmented Fine-tuning for Domain-specific Machine Translation}

\author{
Jiawei Zheng, Hanghai Hong, Feiyan Liu, Xiaoli Wang\thanks{Corresponding Author}, Jingsong Su \\
School of Informatics, Xiamen University \\
\texttt{\{zhengjiawei, hanghaih, feiyanliu\}@stu.xmu.edu.cn},\\ 
\texttt{\{xlwang, jssu\}@xmu.edu.cn} 
\AND
Yonggui Liang, Shikai Wu \\
xFusion Digital Technologies Co., Ltd \\
\texttt{\{liangyonggui, wushikai\}@xfusion.com}
}

\begin{document}
\maketitle
\begin{abstract}
Large language models (LLMs) have shown great potential in domain-specific machine translation (MT). However, one major issue is that LLMs pre-trained on general domain corpus might not generalize well to specific domains due to the lack of domain-specific knowledge. To address this issue, this paper focuses on enhancing the domain-specific MT capability of LLMs, by providing high-quality training datasets and proposing a novel fine-tuning framework denoted by \textbf{\textit{DragFT}}. DragFT augments LLMs via three techniques: ($i$) \textbf{\textit{Dictionary-enhanced prompting}} integrates dictionary information into prompts to improve the translation of domain-specific terminology.; ($ii$) \textbf{\textit{RAG-based few-shot example selection}} provides high-quality examples that simulate both the domain and style characteristics; ($iii$) \textit{\textbf{Fine-tuning with few-shot examples}} further enhances performance when using in-domain examples. We deploy DragFT on three well-known LLM backbones with 13B training parameters to validate its effectiveness. The results on three domain-specific datasets show that DragFT achieves a significant performance boost and shows superior performance compared to advanced models such as GPT-3.5 and GPT-4o. The drastic performance improvement of DragFT over existing LLMs can be attributed to incorporating relevant knowledge while mitigating noise. 
\end{abstract}

\section{Introduction}
Although Large language models (LLMs) have demonstrated remarkable performance in MT, they often fall short of the performance achieved by domain-specific models. To improve the domain-specific machine translation (MT) capability of LLMs, existing works fall into two groups. The first group employs in-context learning (ICL) by feeding LLMs with in-domain translation examples as a demonstration without further fine-tuning~\cite{aycock2024topic, vilar2023prompting, moslemetal2023adaptive, Biao2023}. ICL provides in-context examples that help the model quickly adapt to specific domains and styles. However, its performance depends heavily on the quality and relevance of examples. Another group fine-tunes LLMs with translation instructions to improve the domain-specific MT capability~\cite{wei2022finetuned,moslem2023finetuning}. However, it often requires high computational costs for extra training on specific domains and may weaken the general MT capabilities in LLMs due to over-specialization~\cite{alves-etal-2023-steering}. Therefore, improving the domain-specific MT capability of general-purpose LLMs remains a challenge. First, current systems still struggle with terminology translation. Even domain-adapted models have difficulty with accurately translating domain-specific terminology~\cite{sato-etal-2020-vocabulary}. Second, high-quality in-domain parallel datasets are often required for fine-tuning LLMs.

This paper addresses the above challenges by boosting fine-tuning with few-shot examples to leverage both ICL and fine-tuning benefits. We propose a novel fine-tuning framework, denoted as \textbf{\textit{DragFT}} (\textbf{\textit{D}}ictionary and \textbf{\textit{r}}etrieval \textbf{\textit{a}}u\textbf{\textit{g}}mented \textbf{\textit{F}}ine-\textbf{\textit{T}}uning), to augment the performance of LLMs in domain-specific MT. DragFT contains three components: dictionary-enhanced prompting, RAG-based few-shot example selection, and fine-tuning with few-shot examples. We propose \textbf{\textit{Dict-rephrasing}}, a dictionary-enhanced algorithm, that rephrases the source sentence by replacing terminology with domain-specific terms in the target language. It can augment fine-tuning performance by improving domain-specific terminology translation. A RAG-based few-shot example selection mechanism is developed to boost fine-tuning with high-quality examples in instructions. We use extra corpora (self-constructed domain-specific corpora) to build vector databases and retrieve relevant translation pairs to construct prompts with few-shot examples, which are then fed into LLMs for fine-tuning. To address the scarcity of high-quality parallel translation corpora in specific domains, we construct a domain-specific translation instruction-following dataset in IT domain and enhance the data quality by using LLM-based evaluation and human annotation. Our main contributions are summarized as follows:

\begin{itemize}
\item We propose DragFT, a novel fine-tuning framework designed to enhance domain-specific machine translation. DragFT incorporates dictionary-enhanced prompting to improve terminology translation and utilizes a RAG-based selection mechanism to integrate high-quality examples.

\item We construct a bilingual translation corpus in IT domains and improve data quality through LLM-based evaluation and manual annotation, tackling the challenge of limited high-quality training data for fine-tuning in domain-specific MT.

\item We conduct comprehensive experiments by adapting three well-known 13B backbone models over three datasets in different domains. The results show that DragFT can achieve significant improvements on existing LLMs in domain-specific MT. It also shows superior performance compared with strong baselines.


\end{itemize}

\section{Related Works}

\subsection{ICL in Machine Translation}
ICL feeds LLMs with extra translation examples within the prompts to improve the MT capabilities, without fine-tuning~\cite{Brownllm}. 
Several works focused on improving the MT capabilities of LLMs via ICL.
~\cite{Biao2023} revealed that prompt example effectiveness in MT depends on features like sequence length and semantic similarity, with back-translation being especially robust. \cite{agrawal-etal-2023-context} showed that optimizing in-context examples and prompts, especially using n-gram overlap and re-ranking, significantly improves the MT quality.
Other works investigated prompting strategies for identifying appropriate examples. \cite{vilar2023prompting} evaluated the MT performance of PaLM~\cite{chowdhery2023palm} with different prompting strategies. ~\cite{garcia2022using} used natural language-described prompts to control and improve multilingual MT, enabling translation into specific dialects and unseen languages.~\cite{Jiao2023IsCA} demonstrated that effective prompts and example utilization can enhance ChatGPT~\footnote{\url{https://chat.openai.com}} multilingual translation, with a pivot prompting strategy improving performance for distant languages.

Although moderate progress has been made, ICL is highly sensitive to the quality of provided examples. Poor examples may lead to sub-optimal LLM translation performance.

\subsection{Instruction tuning in Machine Translation}
Instruction tuning is a technique for fine-tuning language models to improve their abilities to follow specific instructions, enhancing their adaptability and performance across diverse downstream tasks. Given labeled domain-specific data, instruction tuning can be an alternative to improve the MT capabilities of LLMs. Instruction tuning is reported to outperform in-context learning in MT performance~\cite{li2023eliciting}.
Several works enhanced the MT performance of LLMs by fine-tuning them with translation instructions on large amounts of parallel data \cite{wei2022finetuned, yang2023bigtrans, zhang2023bayling, chen2023improving}.
\cite{jiao-etal-2023-parrot} incorporated hint fields and three instruction types to enhance chat translations.
\cite{xu2024a} revealed that large parallel datasets are unnecessary for high MT performance in LLMs, achieving significant improvements with a novel two-stage fine-tuning method involving monolingual fine-tuning and lightweight parallel fine-tuning.

\subsection{Domain-specific Machine Translation}
Even though trained on large amounts of data, these two groups of methods can struggle to translate inputs with rare words in domain transfer scenarios~\cite{ghazvininejad2023dictionarybased}. Therefore, several works focused on using the domain-specific vocabulary to supply translations in low-resource settings \cite{lu2023chain,ghazvininejad2023dictionarybased,moslem-etal-2023-domain}. For instance, \cite{ghazvininejad2023dictionarybased} incorporated the additional dictionaries into zero-shot examples without training.

Our work takes full advantage of ICL and instruction tuning, incorporating high-quality and relevant translation examples during the fine-tuning stage. We introduce a RAG-based method for providing high-quality in-domain examples, ensuring the selected examples are semantically similar and contextually relevant to the training data. Additionally, we propose a novel dictionary augmentation method to address the challenge of translating terminology in specific domains.

\begin{figure*}[ht]
    \centering
    \includegraphics[width=\linewidth]{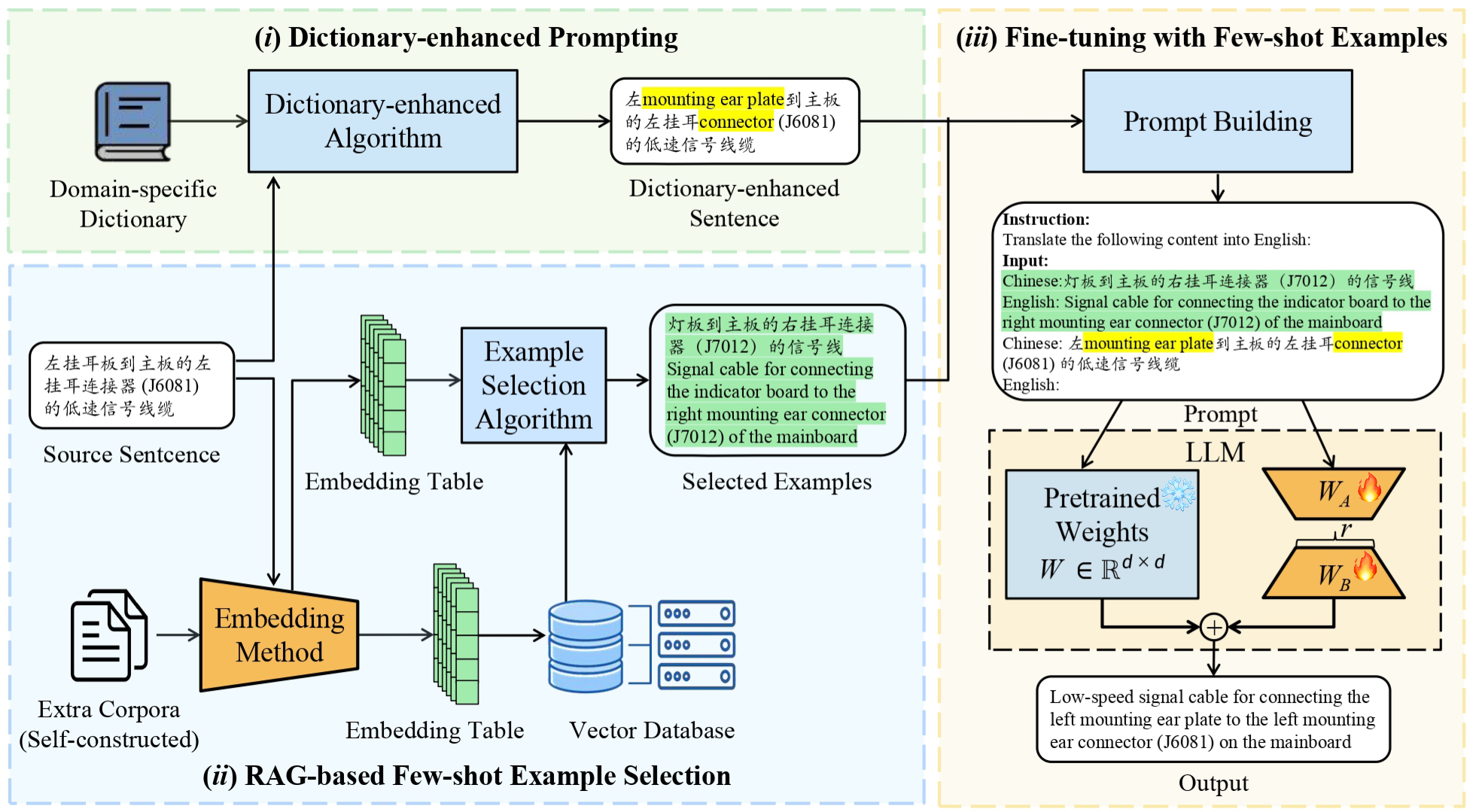}
    \caption{The framework of DragFT, including three techniques: ($i$) \textbf{\textit{Dictionary-enhanced prompting}}, ($ii$) \textbf{\textit{RAG-based few-shot example selection}}, and ($iii$) \textit{\textbf{Fine-tuning with few-shot examples}}.}
    \label{figs:DragFT}
\end{figure*}

\section{DragFT}
As shown in Figure \ref{figs:DragFT}, our DragFT enhances the domain-specific MT capabilities of LLMs through three techniques: ($i$) \textbf{\textit{Dictionary-enhanced prompting}} is a dictionary augmented technique for improving domain-specific terminology translation; ($ii$) \textbf{\textit{RAG-based few-shot example selection}} provides selected examples that closely match the source sentence in both translation style and vocabulary; ($iii$) \textit{\textbf{Fine-tuning with few-shot examples}} incorporates in-domain examples into fine-tuning by taking advantages of both ICL and fine-tuning.

\subsection{Machine Translation Task}
Fine-tuning LLM for adaptation to domain-specific MT requires the guidance of translation instructions. Given a bilingual training dataset of $\mathbf{C}$, which contains pairs of parallel bilingual training data denoted as $(\mathbf{x}, \mathbf{y})$, the optimization function $\mathcal{L}$ for the MT task is defined as follows:

\begin{equation}
  \label{eq:example}
  \mathcal{L} = \sum_{(\mathbf{x}, \mathbf{y}) \in \mathbf{C}} - \log p(\mathbf{y} \lvert \mathbf{x}, \mathcal{T}; \theta)
\end{equation}
where $\mathbf{x}=\{x_1, ..., x_n\}$ is the source sentence, $\mathbf{y}=\{y_1, ..., y_m\}$ is its corresponding target translation, $\mathcal{T}$ is the translation instruction template, and $\theta$ represents the training parameters. The probability of a target sentence given the source sentence is:

\begin{equation}
  \label{eq:example2}
    p(\mathbf{y} \lvert \mathbf{x}, \mathcal{T}; \theta)=\prod_{t=1}^m p\left(y_t \lvert y_{<t}, \mathbf{x}, \mathcal{T}; \theta\right)
\end{equation}
where $y_t$ is the $t$-th generated token, $y_{<t}$ is the privious tokens.



\subsection{Dictionary-enhanced Prompting}
\begin{figure*}[t]
    \centering
    \includegraphics[width=\linewidth]{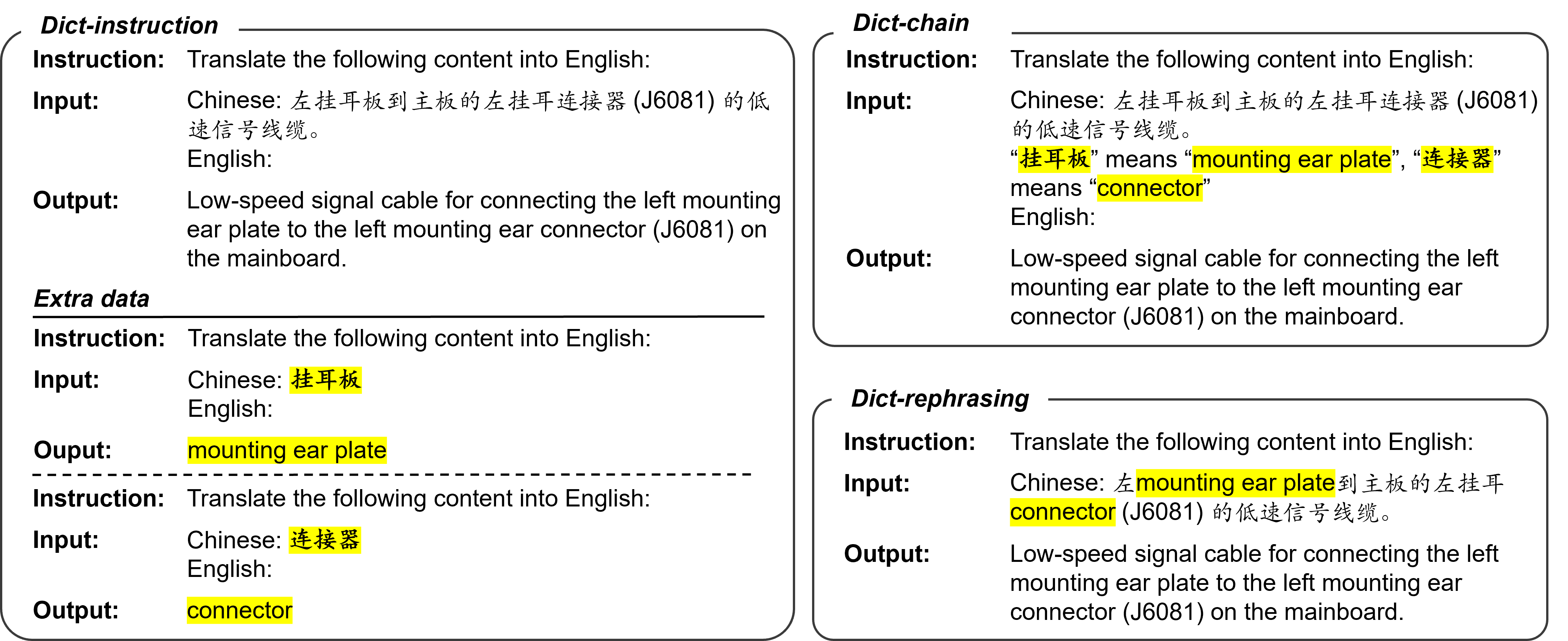}
    \caption{An illustration of three dictionary enhancement prompts, including Dict-instruction, Dict-chain, and Dict-rephrasing.}
    \label{figs:dict}
\end{figure*}

The main obstacle in domain-specific MT lies in the domain-specific terminology that is not commonly used in general domains, which results in inaccurate translations. To tackle this challenge, incorporating domain-specific terminology dictionaries into translation prompts is crucial. One straightforward method combines dictionary data along with the parallel corpus data to create a translation instruction format, called by \textit{\textbf{Dict-instruction}}. Inspired by \cite{Biao2023}, another approach appends the dictionary translation after the sentence translation in a chained manner, named as \textbf{\textit{Dict-chain}}. However, the Dict-instruction increases the amount of fine-tuning data, while the Dict-chain extends the length of prompts, resulting in higher consumption of training resources and longer training time. 

In this paper, we introduce a novel dictionary enhancement algorithm, denoted as \textbf{\textit{Dict-rephrasing}}. It directly replaces the domain-specific terminology in source sentences with their corresponding terms in the target language from the in-domain dictionary,  as illustrated in Algorithm~\ref{dict}. Figure~\ref{figs:dict} shows examples of the three dictionary-enhanced prompting methods. Using the Dict-rephrasing, the terminology of 
``\begin{CJK}{UTF8}{gkai}挂耳板\end{CJK}'' and ``\begin{CJK}{UTF8}{gkai}连接器\end{CJK}'' in the source sentence of ``\begin{CJK}{UTF8}{gkai}左挂耳板到主板的左挂耳连接器(J6081)的低速信号线缆\end{CJK}'' are directly rephrased to ``mounting ear plate'' and ``connector'', respectively. Therefore, the source sentence is rephrased as ``\begin{CJK}{UTF8}{gkai}左mounting ear plate到主板的左挂耳 connector(J6081)的低速信号线缆\end{CJK}''.

\begin{algorithm}[t]
    \caption{Dict-rephrasing}
        {\bf Input:} domain-specific dictionary $\mathcal{D}$, domain-specific parallel corpus $\mathbf{C}$  \\
        {\bf Output:} dictionary-enhanced parallel corpus $\mathbf{C}'$\\
    \begin{algorithmic}
    \STATE \textbf{Sort} $\mathcal{D}$ by length $\downarrow$
    
    \FOR{each translation pair $(x, y)$ in $\mathbf{C}$}
        \STATE Initialize $x' \gets x$
        \FOR{each word pair $(w_{src}, w_{tgt})$ in $\mathcal{D}$}
            \IF{$w_{src}$ in $x$}
                \STATE Replace $w_{src}$ in $x'$ with $w_{tgt}$
                \STATE $x' \gets \text{Replace}(x', w_{src}, w_{tgt})$
            \ENDIF
        \ENDFOR
        \STATE $\mathbf{C}' \gets \mathbf{C}' \cup \{(x',y)\}$
    \ENDFOR
\end{algorithmic}
\label{dict}
\end{algorithm}

Dict-rephrasing helps LLMs better understand the terminology in context, effectively reducing the volume of training data compared to Dict-instruction and shortening the length of prompts compared to the Dict-chain. Our experiments in section~\ref{ablation_study} will further explore the effects of these methods. 

\subsection{RAG-based Few-shot Example Selection}\label{rag-based section}
The main idea of Retrieval-Augmented Generation (RAG)~\cite{lewis2020retrieval} is integrating information from external data sources to supplement the input query or enhance the output. To ensure the quality of few-shot examples, we apply the idea of RAG and design a few-shot example selection mechanism based on it. Specifically, we vectorize extra corpora using the BGE model~\cite{xiao2023c} and store these vectors to construct a domain-specific vector database of $V$. Given a source sentence, we convert it into a vector of $s$ using the BGE model. To retrieve semantically similar and contextually relevant examples from $V$, we calculate the similarity score of $c_i$ between $s$ and the vector of $v_i \in V$.

\begin{equation}
    \label{eq:example3}
    c_i = \frac{s \cdot v_i}{\|s\| \|v_i\|}
\end{equation}
where \( \cdot \) represents the dot product function.

We set a similarity score threshold of $k$ and a maximum number of examples $n$ to refine the selection process. If the similarity score of $c_i$ is greater than $k$, $v_i$ is selected and added to the relevant examples set of $R$. When $\lvert R \lvert$ is equal to $n$, we stop retrieving to limit the volume of the fine-tuning dataset.


\subsection{Fine-tuning with Few-shot Examples} 
We utilize the training dataset with few-shot translation examples to fine-tune LLMs. It is reported that fine-tuning with few-shot examples helps maintain the few-shot learning capabilities of LLMs while preserving the benefits of fine-tuning \cite{alves-etal-2023-steering}. The prompt example adopted in our study is shown in Figure~\ref{figs:DragFT}. We use ``\textit{Translating the following content into <target-language>}'' as the translation instruction with selected examples and sentences to be translated as inputs.
To reduce training costs, we utilize the LoRA~\cite{hu2021lora} fine-tuning strategy, which is designed for efficient fine-tuning of LLMs. As illustrated in Figure~\ref{figs:DragFT}, the pre-trained weights of $W \in \mathbb{R}^{d \times d}$ are frozen, while two low-rank matrices of $W_A$ and $W_B$ with the rank of $r$ are introduced to capture the parameter updates. This approach allows for efficient fine-tuning with reduced computational costs and GPU memory requirements.
\section{Experimental Setups}

\subsection{Datasets}
We conduct experiments with four translation pairs of English to German (en$\rightarrow$de), German to English (de$\rightarrow$en), English to Chinese (en$\rightarrow$zh), and Chinese to English (zh$\rightarrow$en).


For pairs involving zh, we collect documents within the IT domain in both Chinese and English from well-known IT companies, and segment them into sentences to form a parallel IT domain parallel corpus. To improve data quality, we utilize the COMETKiwi~\cite{rei2023scaling}, a model-based evaluation method that doesn't require corresponding translation references. Translation pairs with COMETKiwi scores below $80$ are discarded and the remaining pairs are verified with manual annotations by domain experts.

For pairs involving de, we collect two domains, including Law and Medical from the public corpus Multi-Domain datasets released by ~\cite{aharoni2020unsupervised}.


To generate domain-specific dictionaries, we design prompts for GPT-3.5 to extract terminologies from the training sets. We then work with experts to manually filter out general words and annotate the translations. Detailed prompts are provided in the appendix.

\subsection{Baselines}
To investigate the effectiveness of DragFT, we adapt it on three 13B parameter-scale LLM backbones: \textbf{\textit{Tigerbot-13B}}~\cite{chen2023tigerbot}, \textbf{\textit{Baichuan2-13B}}~\cite{yang2023baichuan}, and \textbf{\textit{LLama2-13B}}~\cite{touvron2023llama}. 
We also consider three well-known strong baselines, including NLLB~\cite{costa2022no} from the NMT domain, GPT-3.5\footnote{The GPT-3.5 version is gpt-3.5-turbo-1106.} and GPT-4o from the LLM domain.

\subsection{Implementation Details}
We fine-tune the backbone models using a learning rate of 3e-4, a training batch size of 2, a maximum sequence length of 512 tokens, a weight decay of 0.00001, and a warm-up ratio of 0.01. For efficient training, we employ the Deepspeed\footnote{\url{https://github.com/microsoft/DeepSpeed}} and Flash-Attention~\cite{dao2022flashattention} acceleration frameworks for fine-tuning with LoRA, with the rank set to 16. In the RAG-based few-shot example selection mechanism, we set the similarity score threshold $k$ to 0.7, and the maximum number of examples $n$ to 2. 
All experiments were conducted on one NVIDIA A100 GPU.

\subsection{Evaluation}
In the inference stage, we adopt the vLLM~\cite{kwon2023efficient} framework to accelerate inference and reduce memory usage. We use the beam search algorithm with a beam width of 4, a temperature of 0 to minimize diversity in translation output, and a length penalty of 1.0.
For translation quality evaluation, we use two widely used evaluation metrics in MT, including the word-based metric of BLEU \cite{papineni2002bleu}, and the reference-based metric of COMET \cite{rei2022comet} for model evaluation.

\begin{table*}[t]
\centering
\resizebox{\textwidth}{!}{
\begin{tabular}{ccccccccccccc}
\toprule
\multirow{3}{*}{\textbf{Model}}    & \multicolumn{6}{c}{\textbf{X $\Rightarrow$ En}} & \multicolumn{6}{c}{\textbf{En $\Rightarrow$ X}} \\
\cmidrule(lr){2-7} \cmidrule(lr){8-13}
 & \multicolumn{2}{c}{\textbf{IT}} & \multicolumn{2}{c}{\textbf{Law}} & \multicolumn{2}{c}{\textbf{Medical}} & \multicolumn{2}{c}{\textbf{IT}} & \multicolumn{2}{c}{\textbf{Law}} & \multicolumn{2}{c}{\textbf{Medical}}\\ 
\cmidrule(lr){2-3} \cmidrule(lr){4-5} \cmidrule(lr){6-7} \cmidrule(lr){8-9} \cmidrule(lr){10-11} \cmidrule(lr){12-13}
 \small & \small BLEU & \small COMET & \small BLEU & \small COMET & \small BLEU & \small COMET & \small BLEU & \small COMET & \small BLEU & \small COMET & \small BLEU & \small COMET \\
\midrule
\multicolumn{13}{c}{\textit{Advanced Models}} \\
NLLB-3.3B &   26.37  &  82.76  &   48.59     &  82.89   &   40.45    & 76.89 &  26.96   &   83.37  &  40.93    &  85.29    &   36.62   &  81.88 \\ 
GPT-3.5 &  29.33   &   84.58   &  36.85   &  83.66  &   40.10     & 82.61 &   34.44  &   85.58  &  31.73    &  84.25   &  35.65    &  82.08  \\ 
GPT-4o &  31.23   &   85.43   &  39.62   &  84.93  &   41.93     &  83.82   &   37.16  &   86.44   &    36.50    &   \textbf{86.10}   &   43.03     &  \textbf{84.98}  \\ 
GPT-4o($w$ DragFT) &  39.79   & \textbf{86.68}   &   46.99  &  85.61  &   \textbf{ 54.22}    &   85.40  &  \textbf{51.84}   &  86.69   &   \textbf{49.67}     &   85.31   &   44.43    &  84.32  \\ 

\midrule
\multicolumn{13}{c}{\textit{Base Model: Tigerbot-13B}} \\
Tigerbot-13B &   25.79  &  82.47  &   32.61    &   81.28   &    36.41       & 81.09 &  27.79  &  82.22 & 25.84  & 82.19  &  27.98 &  81.87 \\ 
DragFT(Tigerbot-13B)    &   45.49  & 85.64   &   \textbf{54.73}    &  \textbf{87.41}  &  49.60 &  85.24 & 45.31 &  86.92 &  48.54 & 84.61  &  47.78 & 84.39 \\ 
\midrule
\multicolumn{13}{c}{\textit{Base Model: Baichuan2-13B}} \\
Baichuan2-13B   &  26.81     & 82.67  &   32.76  &   82.04   &  35.12    &  81.44 &  30.02  &  82.87  &  26.76 & 82.41  & 25.02  & 79.31  \\ 
DragFT(Baichuan2-13B) & 43.24 &  84.65 & 50.27  & 84.79   & 45.88  & 82.68  &  44.56  &  87.05  & 44.74 & 82.16  & 46.75   &  83.78  \\ 
\midrule
\multicolumn{13}{c}{\textit{Base Model: Llama2-13B}} \\
Llama2-13B &   22.21    &   80.36   &  32.34 &  81.72  &  39.09   & 82.47   &  23.31  & 79.56  & 26.85 &  82.20 & 28.96 &80.00\\ 
DragFT(Llama2-13B) & \textbf{45.64}  &  85.55  & 53.88 & 86.82 & 47.64  &  \textbf{85.49}  &  45.16  &  \textbf{87.07}  &  47.96  &  84.67  &  \textbf{48.54} & 84.72\\ 
\bottomrule
\end{tabular}
}
\caption{Translation performance of advanced models and applying DragFT method on three backbone models (TigerBot-13B, Baichuan2-13B, and Llama2-13B) on IT, Law, and Medical domains testsets.}
\label{table:main-new}
\end{table*}

\section{Results}

Table~\ref{table:main-new} presents the main results of domain-specific translation for X$\Leftrightarrow$En, where X represents Zh for the IT domain and De for the Law and Medical domains. To ensure the consistency between training and testing, we apply the corresponding dictionary-enhanced methods to construct the test set during the inference stage. Overall, our DragFT significantly improves the translation quality of existing LLMs and shows superior performance compared with advanced translation models. We have the following observations:

(1) DragFT achieved significant performance improvements across three LLM backbones in the IT, Law, and Medical domain test sets. This indicates that fine-tuning with high-quality parallel data is the most direct and effective method for adapting LLMs to domain-specific translation tasks. 



(2) Among three advanced of GPT-3.5, GPT-4o, and NLLB-3.3B, GPT-4o achieves the best basic translation performance. 

(3) DragFT demonstrates drastic improvement in the BLEU metric compared to the COMET metric. Since BLEU evaluates translation quality at word and phrase levels, our dictionary-enhanced prompting can augment LLMs by translating domain-specific terminologies. This also indicates the effectiveness of Dict-rephrasing.

(4) Notably, we conduct additional experiments using dictionary-enhanced prompts and a RAG-based examples selection on GPT-4o. We observed a significant improvement in translation quality compared to not using the DragFT enhancement. For large models of this scale, our method achieves superior translation performance even without fine-tuning.





\section{Analysis}

\subsection{Effect of Dictionary-enhanced Prompting}\label{effects of dictionary}
\begin{figure}[ht]
    \centering
    \includegraphics[width=\linewidth]{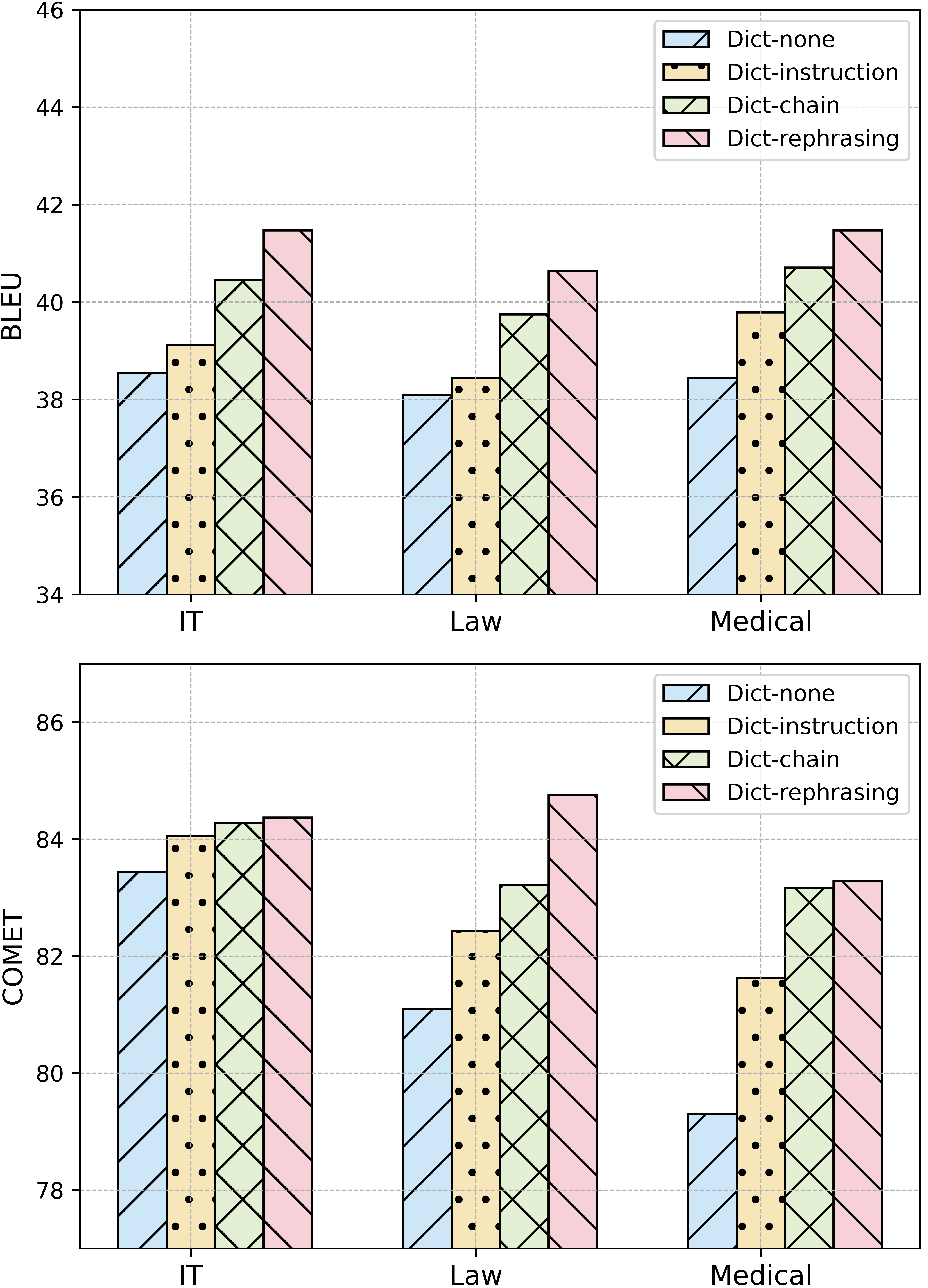}
    \caption{Performance comparison of different dictionary-enhanced prompting methods on domain-specific test sets.} 
    \label{figs:dict_result}
\end{figure}

To investigate whether our proposed dictionary-enhanced algorithm can improve the performance of LLMs in domain-specific MT, we conduct comparative experiments on Tigerbot-13B. We employ three different dictionary-enhanced methods introduced in section~\ref{rag-based section} to construct training data for fine-tuning and then evaluate the translation quality on three domain-specific test sets. We also conduct an experiment on fine-tuning without dictionary augmentation, denoted as \textbf{\textit{Dict-none}}. The experimental results are shown in Figure~\ref{figs:dict_result}.

Compared to Dict-none, all three dictionary-enhanced methods demonstrate translation performance improvements, indicating that they can effectively improve domain-specific terminology translation quality. Among them, our proposed Dict-rephrasing algorithm shows the most significant improvement, although it performs slightly worse than the Dict-chain in the Medical domain. This strongly validates the effectiveness of our proposed Dict-rephrasing, which directly embeds terminology information into the source sentences. This approach neither requires additional dictionary data for training nor increases the prompt length, allowing the LLMs to better understand the context of terminology during training, and therefore improving the translation quality.

\subsection{Ablation Study}\label{ablation_study}
We conduct an ablation study to analyze the effects of different components of DragFT. Table~\ref{table:ablation_fix} shows the results on Tigerbot-13B, which highlights the importance of each component in DragFT. 

\begin{table*}[t]
\centering
\resizebox{0.95\textwidth}{!}{
\begin{tabular}{clcccccc}
\toprule
\multirow{2}{*}{\textbf{ID}}  & \multirow{2}{*}{\textbf{Method}}   & \multicolumn{2}{c}{\textbf{IT}} & \multicolumn{2}{c}{\textbf{Law}} & \multicolumn{2}{c}{\textbf{Medical}} \\ 
\cmidrule(lr){3-4} \cmidrule(lr){5-6} \cmidrule(lr){7-8} 
\small & \small  & \small BLEU & \small COMET & \small BLEU & \small COMET &  \small BLEU & \small COMET  \\
\midrule
\textbf{0} & \textbf{DragFT~[Dict-rephrasing]}   &   \textbf{45.49 } & \textbf{85.64}   &  \textbf{54.73}    &  87.41  &  49.60 &  \textbf{85.24 } \\
\textbf{1} & $~~~~w/o$ Dict-rephrasing & 42.25 & 84.02 & 48.74  & 84.01 & 44.57 & 83.67\\
\textbf{2} & $~~~~w/o$ RAG-based selection  & 39.42  & 80.41  & 50.78 & 85.69 & 45.98 & 84.37 \\ 
\textbf{3} & $~~~~w/o$ few-shot example  & 41.47 & 84.37 & 48.45 & 83.75 & 43.17 & 83.98\\ 
\textbf{4} & DragFT~[Dict-instruction]   & 43.89 & 84.34 & 53.40 & \textbf{87.98} & 46.12 & 84.78\\ 
\textbf{5} & DragFT~[Dict-chain]  & 44.47 & 84.87 & 52.82 &  87.01 & \textbf{49.78} & 85.17\\ 


\bottomrule
\end{tabular}
}
\caption{Ablation study. We report the BLEU and COMET scores in X$\Rightarrow$En direction with Tigerbot-13B.}
\label{table:ablation_fix}
\end{table*}
\begin{itemize}
\item \textbf{\textit{Without (w/o) Dict-rephrasing}}. We remove Dict-rephrasing and use the source sentence. From result IDs of 0 and 1 in Table~\ref{table:ablation_fix}, we observe a significant drop in translation quality without dictionary-enhanced prompting. This indicates its essential role in domain-specific MT. The results of 0, 4, and 5 show that the Dict-rephrasing algorithm achieves superior performance compared to the Dict-instruction and Dict-chain methods, which also validates our findings in section~\ref{effects of dictionary}, indicating the effectiveness of the Dict-rephrasing algorithm for domain-specific MT.

\item\textbf{\textit{Without (w/o) RAG-based selection}}. We replace the RAG-based example selection mechanism with a strategy that randomly selects two examples for each training data from extra corpora. The results of 0 and 2 in Table~\ref{table:ablation_fix} reveal a remarkable performance decline in the LLM without RAG selection, which also indicates the quality and relevance of examples can affect the performance.

\item\textbf{\textit{Without (w/o) few-shot example}}. We directly conduct instruction tuning on the LLM without providing any translation examples. From the results of 0 and 3, we find a drastic decline in translation quality when performing instruction tuning without few-shot examples. This suggests that simple instruction tuning is insufficient to fully leverage the ICL capabilities of LLMs.
\end{itemize}

\subsection{Effects of DragFT}
\begin{table}[h]
\centering
\resizebox{0.5\textwidth}{!}{
\begin{tabular}{lccc}
\toprule
\textbf{Method} & \textbf{IT} & \textbf{Law} & \textbf{Medical} \\
\midrule
\textbf{GPT-4o} & 73.05 & 63.49 & 69.00 \\
\textbf{GPT-4o($w$ DragFT)} & \textbf{83.64} & 72.17 & 78.37 \\
\midrule
\textbf{Tigerbot-13B} & 57.86 & 52.36 & 62.40 \\
\textbf{DragFT(Tigerbot-13B)} & 80.07 & \textbf{75.61} & \textbf{79.67} \\
\bottomrule
\end{tabular}
}
\caption{Success rate of Terminology Translation (\%) on domain test sets(X$\Rightarrow$En).}
\label{table:term}
\end{table}
To analyze the influence of the DragFT framework, we perform the following two comparisons.

\textbf{Success rate of Terminology Translation} To analyze the improvement of our proposed DragFT in domain terminology translation, we reference the metrics from the WMT-23 Terminology Shared Task~\footnote{https://wmt-terminology-task.github.io/}. Table~\ref{table:term} shows that compared to GPT-4o, both the dictionary-enhanced prompt version (GPT-4o $w$ DragFT) and the dictionary-enhanced fine-tuned model (DragFT) significantly improve terminology translation success rates. The fine-tuned model, in particular, shows more substantial improvements, suggesting that fine-tuning is more effective in LLMs to specific domain translation tasks.

\begin{figure}[ht]
    \centering
    \includegraphics[width=\linewidth]{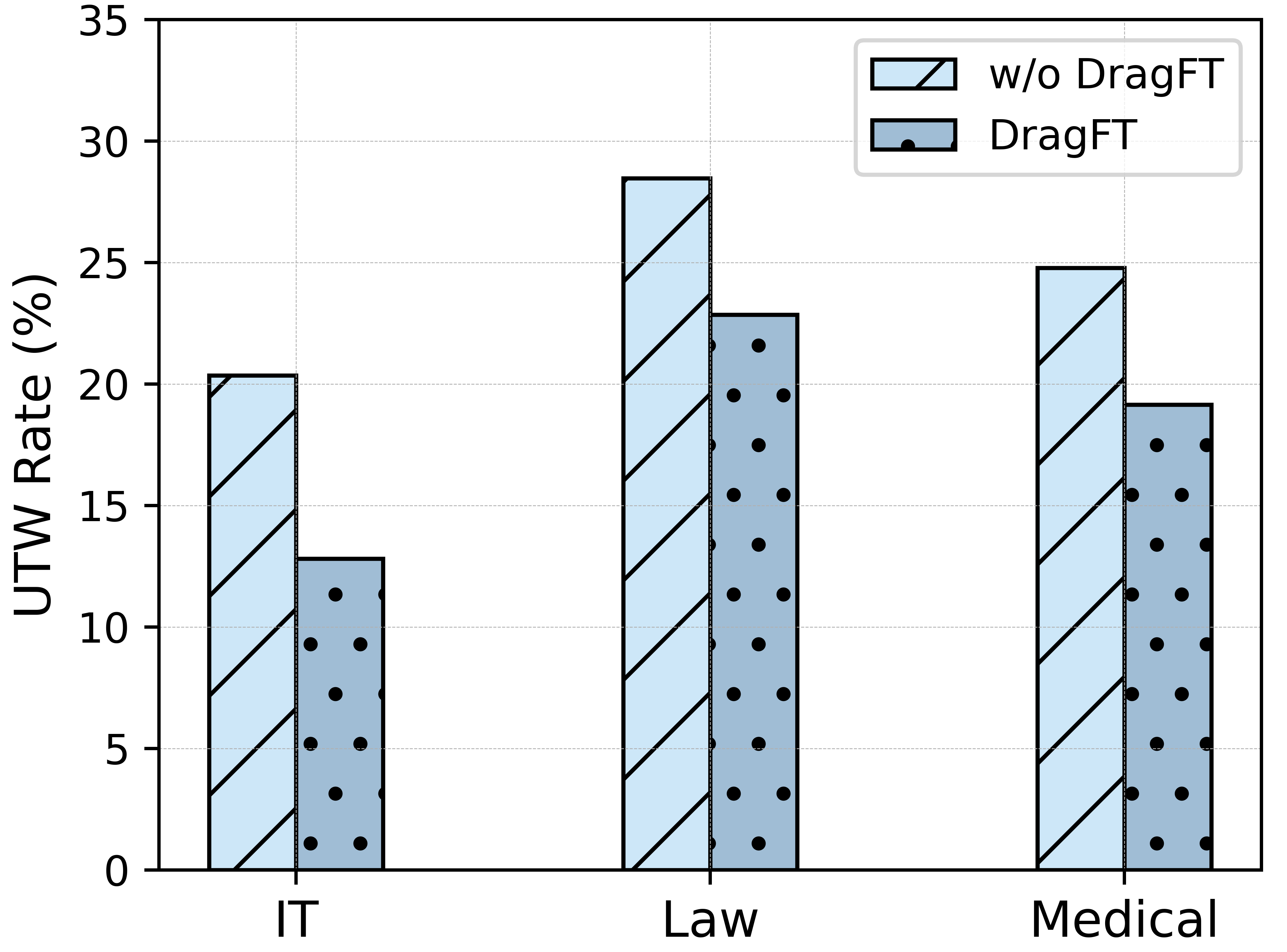}
    \caption{Comparison between the UTW before and after applying DragFT.}
    \label{figs:usw}
\end{figure}

\textbf{Unaligned Translation Words (UTW)}~~~To analyze the impact of the DragFT method, we compare the remaining unaligned in a word-to-word alignment rate between target sentences and reference sentences before and after applying DragFT on Tigerbot-13B. The alignment is measured using the method from \cite{dou2021word}, also used by \cite{hendy2023good}. 
The results are shown in Figure~\ref{figs:usw}, we can observe that after domain adaptation with DragFT, the UTW significantly decreased, indicating improved word translation precision and overall translation performance. This validates DragFT's advantage in handling domain-specific terms.


\section{Conclusion}
To enhance the domain-specific MT capabilities of LLMs, this paper proposes a novel fine-tuning framework denoted as DragFT. DragFT employs dictionary-enhanced prompting to improve domain-specific terminology translation and RAG-based few-shot example selection to provide high-quality few-shot examples to boost fine-tuning with in-domain examples. We deploy DragFT on three well-known LLM backbones, and the results on three domain-specific datasets show that DragFT can achieve a remarkable performance boost in three backbones and surpass strong baselines. The performance improvement of DragFT over existing LLMs can be attributed to the incorporation of relevant knowledge while mitigating noise. We also construct an IT domain translation corpus in Zh$\Leftrightarrow$En to accelerate future research in domain-specific MT. Our current proposed framework fine-tunes all instances, irrespective of whether a test instance requires fine-tuning or not, which may lead to the deterioration of translation quality for some sentences. In the future, we plan to identify those sentences that require fine-tuning and adapt only to them. Meanwhile, we perform dictionary-enhanced prompting for all instances, irrespective of whether a terminology requires enhancement or not, which may lead to the deterioration of translation quality for some sentences. Moving forward, we will focus on identifying domain-specific terms that require rephrasing or dictionary chaining and adopt only those.

\section*{Limitation}
We focus on the Zh$\Leftrightarrow$En and De$\Leftrightarrow$En translation directions and have not validated the effectiveness of our methods on low-resource languages. Due to time and resource constraints, we rely on machine translation metrics rather than human evaluation to assess translation quality.

\section*{Ethics Statement}
This work relies on large language models which, as detailed in~\cite{Brownllm} and~\cite{chowdhery2023palm}, can carry inherent risks. Potential issues include the presence of toxic content due to training on extensive web corpora~\cite{gehman2020realtoxicityprompts}, and high energy consumption during training~\cite{strubell2019energy}. In constructing the domain-specific dataset, the data were collected with respect to individual privacy, and proper consent was obtained where applicable. Personal or sensitive information was anonymized to ensure protection. Furthermore, to enhance the quality of the dataset, we engage annotators who are duly compensated for their time and expertise, ensuring fair practices by established standards.


\bibliography{custom}




\clearpage
\newpage

\appendix

\setcounter{table}{0}   
\setcounter{figure}{0}
\setcounter{section}{0}
\setcounter{equation}{0}
\renewcommand{\thetable}{\arabic{table}}
\renewcommand{\thefigure}{\arabic{figure}}
\renewcommand{\thesubsection}{A\arabic{subsection}}
\renewcommand{\theequation}{\arabic{equation}}

\section*{Appendix A} 
\begin{table*}[ht]
\centering
\resizebox{\textwidth}{!}{
\begin{tabular}{ll}
\toprule

\textbf{Source} & \begin{CJK}{UTF8}{gkai}（适用于\textcolor{blue}{整机柜}服务器/节点，DH141系列除外）当管理网口切换到NC-SI\end{CJK}\\
&\begin{CJK}{UTF8}{gkai}通道时，面板的GE管理网口会变更成近端维护网口\end{CJK}\\
\textbf{Reference} & (For \textcolor{blue}{Rack-Scale} servers/nodes, except DH141 series) When the management \\
&network port is switched to the NC- SI channel, the GE management network port on the \\
&panel is changed to the local maintenance network port.\\
\midrule
\textbf{Tigerbot-13B} & When the management network port switches to the NC-SI channel, the GE\\ &management network port becomes a near-end maintenance port.\\
\textbf{GPT-4o} & (Applicable to \textcolor{red}{full-rack} servers/nodes, except for the DH141 series) When the management\\
&  network port switches to the NC-SI channel, the GE management network port\\
& on the panel will change to a local maintenance port.\\
\textbf{DragFT} & (Applicable to \textcolor{blue}{rack-scale} servers/nodes, except DH141 series) When the management \\
&network port is switched to the NC-SI channel, the GE management network port on the \\
&  panel changes to the local maintenance network port.\\

\bottomrule
\end{tabular}
}
\caption{Case Study}
\label{table:case}
\end{table*}
\subsection{Case Study}
In this case study, we compare the translation outputs from Tigerbot-13B, GPT-4o, and DragFT for an IT domain parallel translation pair(Zh$\Rightarrow$En). The aim is to highlight the strengths of DragFT in terms of accuracy and alignment with the context.
Firstly, it is important to note that the source text involves specialized terminology and nuanced technical information pertinent to server management. This specificity demands high translation fidelity, both in terms of lexical choice and the preservation of meaning.
The reference translation provided sets a benchmark for evaluating the outputs, emphasizing the importance of correct term usage, such as "Rack-Scale" for \begin{CJK}{UTF8}{gkai}整机柜\end{CJK} and "local maintenance network port" for \begin{CJK}{UTF8}{gkai}近端维护网口\end{CJK}.
Tigerbot-13B's translation, while generally understandable, loses technical precision by omitting specific terms like "Rack-Scale" and simply referring to a "near-end maintenance port" without context. This could lead to ambiguity in a technical document.
In contrast, GPT-4o introduces a notable discrepancy by translating \begin{CJK}{UTF8}{gkai}整机柜\end{CJK} as "full-rack" instead of "Rack-Scale." This substitution, although semantically related, deviates from the standard terminology often used in server management contexts, potentially leading to misunderstandings among professionals in the field.
DragFT, however, delivers a highly accurate and contextually faithful translation. It correctly translates \begin{CJK}{UTF8}{gkai}整机柜\end{CJK} as "Rack-Scale," aligning perfectly with industry terminology. Furthermore, DragFT retains the structure and technical nuance of the original text, ensuring that "the GE management network port on the panel changes to the local maintenance network port" is clearly and accurately conveyed. This adherence to both lexical accuracy and contextual integrity makes DragFT stand out as the superior translation model in this scenario.

\subsection{Domain-specific Dictionary Generation}

We employ a method combining LLM models and manual annotation to build domain-specific dictionary data. The process is outlined as follows:

\begin{enumerate}
    \item For three domain-specific datasets (IT, Law, Medical), we initially input data into GPT-3.5 using predefined prompts, as shown in Table~\ref{table:build}.
    \item The LLM model extracts domain-specific words from the data guided by the prompts.
    \item Domain experts perform manual annotations to enhance the accuracy of translating specialized terms.
\end{enumerate}

This approach integrates automated text processing capabilities with domain expertise from human professionals, enabling the efficient generation of high-quality and precise domain-specific dictionary data.

\begin{table*}[ht]
\centering
\resizebox{\textwidth}{!}{
\begin{tabular}{ll}
\toprule

\textbf{Prompt} & You are a seasoned translator specializing in the IT domain. Please review \\
&the provided Chinese-English translation pairs and indentify the most specialized\\
& IT terms from each pair.Skip any pairs that do not contain specialized IT terms.\\

\textbf{Input} & Chinese:\begin{CJK}{UTF8}{gkai}在每个\textcolor{blue}{元数据服务器}上执行如下命令查询MDS\textcolor{blue}{数据盘}\textcolor{blue}{使用量}。\end{CJK}\\
& English:Run the following command on each \textcolor{red}{metadata server} to query the \\
& MDS \textcolor{red}{data disk usage}.\\

\textbf{Output} & \textcolor{blue}{\begin{CJK}{UTF8}{gkai}元数据服务器\end{CJK}}    \textcolor{red}{metadata server}\\ 
& \textcolor{blue}{\begin{CJK}{UTF8}{gkai}数据盘\end{CJK}} \textcolor{red}{data disk}\\
& \textcolor{blue}{\begin{CJK}{UTF8}{gkai}使用量\end{CJK}} \textcolor{red}{usage} \\

\bottomrule
\end{tabular}
}
\caption{The Prompt for GPT-3.5 to construct domain-specific dictionary data.}
\label{table:build}
\end{table*} 

\subsection{Effect of Instruction Tuning on MT}
To evaluate the effect of instruction tuning on MT tasks, we conduct a comparative experiment using the Tigerbot-13B. We use the WMT22 test set (Zh$\Leftrightarrow$En)~\footnote{\url{https://www.statmt.org/wmt22}} as the test set, which is formatted into translation instructions. Additionally, we extract 20,000 samples from the WMT19 parallel corpus (Zh$\Leftrightarrow$En) to form the training set.

The experiment includes the following settings:





\textbf{Pre-trained}: The test set is directly fed into the original model without fine-tuning.

\textbf{Fine-tuned}: The model is fine-tuned using training data without translation instruction tuning.

\textbf{Instruction-tuned}: The model is fine-tuned using training data formatted with translation instructions.

\textbf{Reference}: The referenced translations of the test set.

\begin{figure}[ht]
    \centering
    \includegraphics[width=\linewidth]{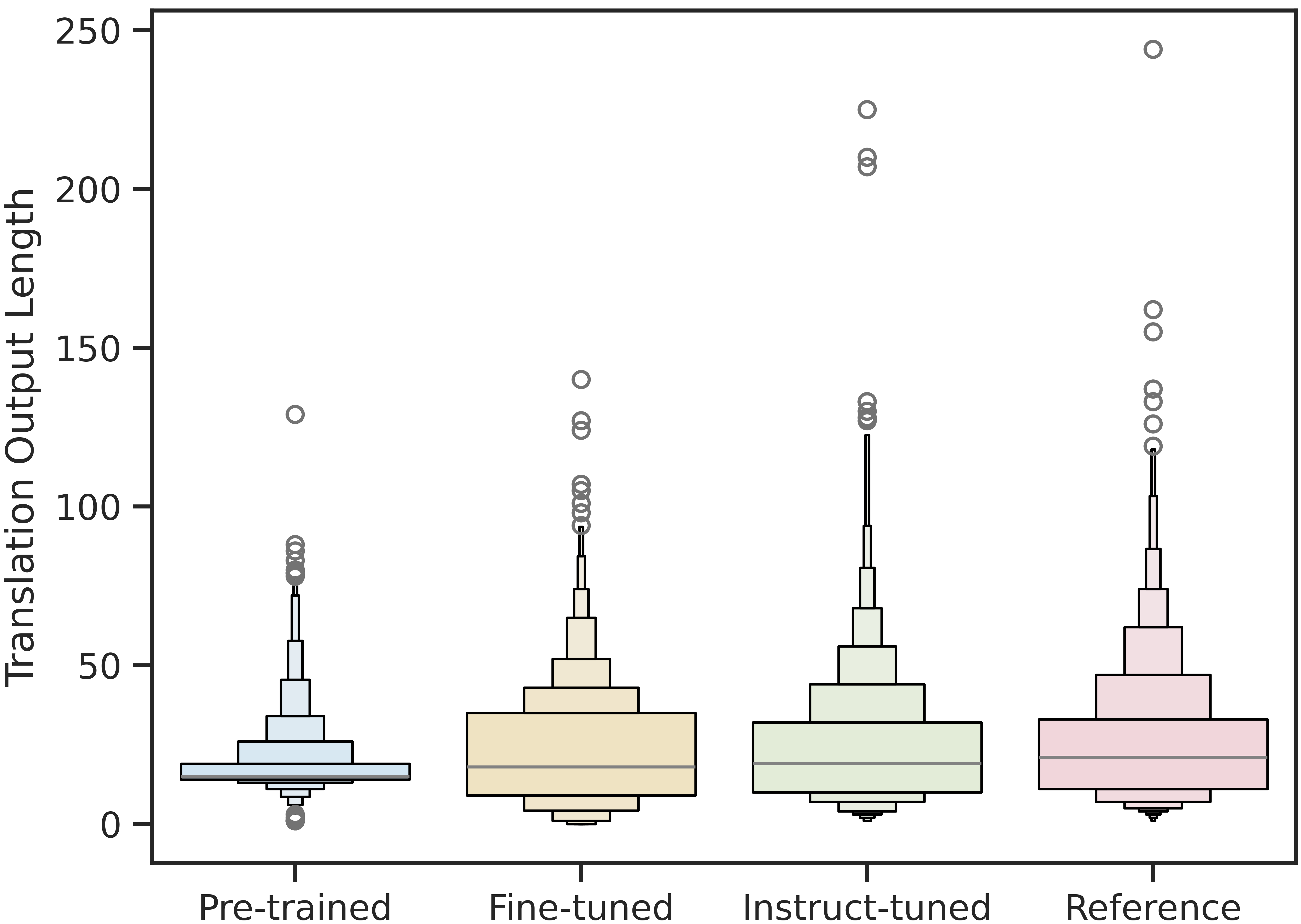}
    \caption{The length distribution of tokenized outputs on the WMT22 test set (Zh$\Rightarrow$En).}
    \label{figs:length}
\end{figure}

We show the length distribution result of tokenized outputs when translating the WMT22 test set (Zh$\Rightarrow$En) on different training setups as shown in Figure~\ref{figs:length}. We observe that the outputs of the pre-trained model are generally too short, indicating a failure to accurately understand the MT task without fine-tuning. On the other hand, the fine-tuned model produces excessively long outputs, demonstrating the over-generation problem. In contrast, the instruction-tuned model generates outputs with length distribution closer to the reference. This indicates that instruction tuning effectively guides the model to complete the MT task without generating redundant information.

\subsection{Dataset Statistics}
After separating the test set, we select 60,000 manually screened, high-quality bilingual parallel data for fine-tuning in each of the three domains (IT, Law, and Medical). The remaining data is used to build the vector database. Table~\ref{tab:data_statics} shows the statics of the datasets we construct on three specific domains.
\begin{table}[ht]
    \centering
    \begin{tabular}{lccc}
        \toprule
        \textbf{Domain} & \textbf{Train} & \textbf{Test} & \textbf{Vector Database} \\
        \midrule
        IT       & 60k & 4.9k & 75k \\
        Law      & 60k  & 2.0k & 100k \\
        Medical  & 60k  & 2.0k & 87k \\
        \bottomrule
    \end{tabular}
    \caption{The data statistics of the datasets we construct on three domain-specific datasets.}
    \label{tab:data_statics}
\end{table}
\begin{table}[ht]
\centering
\resizebox{0.5\textwidth}{!}{
\begin{tabular}{lccc}
\toprule
\textbf{Method} & \textbf{IT} & \textbf{Law} & \textbf{Medical} \\
\midrule
\textbf{Dict-chain} & 1.54M & 4.87M & 4.10M \\
\textbf{Dict-rephrasing} & 1.24M & 2.66M & 2.02M \\
\bottomrule
\end{tabular}
}
\caption{Length of token using different dictionary enhancement methods.}
\label{tab:length1}
\end{table}

\begin{table}[ht]
\centering
\resizebox{0.5\textwidth}{!}{
\begin{tabular}{lccc}
\toprule
\textbf{Method} & \textbf{IT} & \textbf{Law} & \textbf{Medical} \\
\midrule
\textbf{Dict-instruction} & 64k & 88k & 75k \\
\textbf{Dict-rephrasing} & 60k & 60k & 60k \\
\bottomrule
\end{tabular}
}
\caption{The number of training data using different dictionary enhancement methods.}
\label{tab:length2}
\end{table}
\begin{table*}[t]
\centering
\resizebox{\textwidth}{!}{
\begin{tabular}{lcccccccc}
\toprule
\multirow{3}{*}{\textbf{Method}}   & \multicolumn{4}{c}{\textbf{Zh $\Rightarrow$ En}} & \multicolumn{4}{c}{\textbf{En $\Rightarrow$ Zh}}\\ 
\cmidrule(lr){2-5} \cmidrule(lr){6-9}
& \multicolumn{2}{c}{\textbf{WMT22}} & \multicolumn{2}{c}{\textbf{Flores-200}}   & \multicolumn{2}{c}{\textbf{WMT22}} & \multicolumn{2}{c}{\textbf{Flores-200}}\\ 
\cmidrule(lr){2-3} \cmidrule(lr){4-5} \cmidrule(lr){6-7} \cmidrule(lr){8-9}  
& BLEU &  COMET & BLEU &  COMET & BLEU &  COMET & BLEU &  COMET  \\ \midrule

\textbf{WMT22 Winners} &  \textbf{33.50} & \textbf{81.0} & - & - & \textbf{54.3} & \textbf{86.8} & - & -\\

\textbf{NLLB-3.3B} & 21.07 & 76.92 & \textbf{29.54}& 86.14  & 32.52& 81.56 & 26.94 & 78.03\\
\midrule
\textbf{Tigerbot-13B} & 15.72  &  76.62  & 27.28 & 86.65 &  36.34 &  85.35  & \textbf{39.89}  &  \textbf{86.94}\\
\textbf{DragFT} & 23.23  &  79.93  & 27.79 & \textbf{86.79} &  40.31 &  86.38  & 38.91  & 86.59 \\

\bottomrule
\end{tabular}
}
\caption{Translation performance of our DragMT on WMT22 test set and Flores-200 test set with Tigetbot-13B model.}
\label{table:general_result}
\end{table*}

\subsection{Benefits of Dict-rephrasing}
We apply three dictionary enhancement methods and conduct data statistics on three training sets. Table~\ref{tab:length1} shows the total token length of instructions and inputs, while Table~\ref{tab:length2} displays the number of training data. It can be observed that compared to the Dict-chain method, the training set enhanced by the Dict-rephrasing has a reduced total token length. In comparison to the Dict-instruction method, Dict-rephrasing significantly reduces the volume of training data. Overall, the Dict-rephrasing method effectively shortens training time by reducing prompt length and data scale, saving time and computational resources.

\subsection{Translation performance in general domain}
To validate the performance of the model fine-tuned with DragFT in the general domain, we evaluate translation metrics on the WMT22 and Flores-200 test sets and compare them with advanced models. The backbone model is Tigerbot-13B. Table ~\ref{table:general_result} shows the results in the general domain. It is evident that DragFT maintains robust domain-specific translation capabilities while demonstrating excellent translation performance on general domain datasets WMT22 and Flores-200~\cite{costa2022no}.

\end{document}